\PassOptionsToPackage{unicode}{hyperref}
\PassOptionsToPackage{hyphens}{url}
\documentclass[
]{article}
\usepackage{xcolor}
\usepackage{amsmath,amssymb}
\usepackage{graphicx}
\usepackage{iftex}
\ifPDFTeX
  \usepackage[T1]{fontenc}
  \usepackage[utf8]{inputenc}
  \usepackage{textcomp} 
\else 
  \usepackage{unicode-math} 
  \defaultfontfeatures{Scale=MatchLowercase}
  \defaultfontfeatures[\rmfamily]{Ligatures=TeX,Scale=1}
\fi
\usepackage{lmodern}
\ifPDFTeX\else
\fi
\IfFileExists{upquote.sty}{\usepackage{upquote}}{}
\IfFileExists{microtype.sty}{
  \usepackage[]{microtype}
  \UseMicrotypeSet[protrusion]{basicmath} 
}{}
\makeatletter
\@ifundefined{KOMAClassName}{
  \IfFileExists{parskip.sty}{%
    \usepackage{parskip}
  }{
    \setlength{\parindent}{0pt}
    \setlength{\parskip}{6pt plus 2pt minus 1pt}}
}{
  \KOMAoptions{parskip=half}}
\makeatother
\usepackage{longtable,booktabs,array}
\usepackage{calc} 
\usepackage{etoolbox}
\makeatletter
\patchcmd\longtable{\par}{\if@noskipsec\mbox{}\fi\par}{}{}
\makeatother
\IfFileExists{footnotehyper.sty}{\usepackage{footnotehyper}}{\usepackage{footnote}}
\makesavenoteenv{longtable}
\ifLuaTeX
  \usepackage{luacolor}
  \usepackage[soul]{lua-ul}
\else
  \usepackage{soul}
\fi
\usepackage{bookmark}
\IfFileExists{xurl.sty}{\usepackage{xurl}}{} 
\hypersetup{
  hidelinks,
  pdfcreator={LaTeX via pandoc}}

\author{}
\date{}

\begin{document}

\section{\textbf{Marking Contour Tones in Yorùbá:}}

\textbf{A Typographic and Computational Proposal}

\emph{By Kọ́lá Túbọ̀sún}

Yoruba Names Project (project@yorubaname.com \textbar{}
 www.YorubaName.com) 
 \vspace{6pt}

\section{Abstract}\label{abstract}

\emph{Yorùbá is a tonal language in which contour tones (two distinct
tones housed within one single syllable) pose persistent orthographic
challenges. These are especially notable for personal names and lexical
items whose conventional spellings avoid vowel lengthening that would
otherwise provide a host syllable for the second tone. A particular
concern is a class of names in which the conventional spelling does not
just omit tonal information but inverts the
meaning of said name, sometimes asserting the opposite of what the name
intends. This paper describes the problem, illustrates the inadequacy of
current solutions, and proposes the adoption of the caron (ˇ) and
circumflex (ˆ) marks. These are symbols with precedent in Yorùbá
phonological scholarship since Olmsted (1951), used as orthographic
conventions on single vowels to encode rising and falling contour tones,
making them accessible for the first time through standard keyboard
input and computational text processing. The proposal is supported by an implementation in the WriteYoruba keyboard and the TTSYoruba speech synthesizer, whose architecture and listener evaluation are reported separately (Túbọ̀sún, Olúòkun, et al., 2026).}

\emph{\textbf{Keywords:} Yorùbá orthography, contour tones, diacritics,
natural language processing, language technology, personal names}

\section{1. Introduction}\label{introduction}

Yorùbá's three tones (High, Mid, and Low) are phonemically contrastive.
They are conventionally marked in standard orthography with acute (´),
macron (¯) or absence of mark, and grave (\textasciigrave) accents respectively on
vowel letters (Capo, 1989).

High tones are the most stable under phonological pressure, Low tones
are intermediate, while Mid tones are the most unstable. Mid tone is the
most readily lost to deletion and epenthesis, and the one unable to
participate in lexical contour formation (Pulleyblank 2004; Akinlabí
1985; Oyěláràn 1973)\footnote{The names in this paper are written with
  the caron and circumflex notation proposed here, which may differ from
  the spellings used by the name-bearers themselves or in prior
  publications. This reflects the paper\textquotesingle s argument that
  the conventional spellings are orthographically incomplete; the
  contour-marked forms used here represent the names\textquotesingle{}
  full tonal structure as the notation is intended to capture it.}.
Empirical evidence for this hierarchy in L2 acquisition contexts is
provided in Ọlátúbọ̀sún (2012), where an elicited imitation study with
American English speakers showed that contour tone errors were common in
H and L environments, with Mid tone the least reliably produced. This
explains the proposal advanced here (\emph{see Section~3}) that the
contour tones requiring orthographic representation are exclusively L+H
(rising) and H+L (falling) sequences. The absence of contour tones
involving Mid tone from this corpus reflects a structural property of
Yorùbá phonology.

Contour tones, which occur when two tonemes are realized within a single
syllable, arise frequently in Yorùbá through phonological processes
including vowel elision, assimilation, epenthesis, and historical
compounding. In fully-marked writing, these are typically resolved by
inserting a doubled vowel so that each tone may occupy its own syllable.
For instance, the English word ``ball'' becomes ``bọ́ọ̀lù'' in Yorùbá. The
personal name ``Nikẹ'' (``one who is ennobled with care/cherishing'')
carries a falling-then-rising contour on its final syllable when spoken,
thus mandating it to be written with a duplicated vowel as Níkẹ̀ẹ́, were
its component tones written out in full.\footnote{It will be excusable,
  tonologically and morphologically, to \emph{write} the name as
  ``Níìkẹ́'', since the word is a combination of ``Ní''(to have) and
  ``Ìkẹ́'' (cherishing/care), but it's not the most tonally or visually
  intuitive for most visual speakers, hence ``Níkẹ̀ẹ́'' is used here; but
  the consequences for contour are the same.}

In practice, however, no one writes the name as ``Níkẹ̀ẹ́'' (though
``bọ́ọ̀lù'' is common). A large number of personal names typically omit
this vowel doubling, either because of familiarity with the current
spelling, or for other personal, historical, or aesthetic reasons. This
creates a structural problem for the writer or editor/publisher who
insists on writing the names correctly with appropriate tone marks
without either altering the accepted spelling or leaving the contour
tone unrepresented. In a second category described below, the meaning of
some names are inverted when the contour tone on a key syllable is
obscured by a convenient improvised spelling, where the acute or grave
accent is used, suppressing the visual representation of the negation
morpheme. A name currently written as ``Fájuyì'', for instance, has
``Fáàjuyì'' as its fuller tonal form.

We propose the caron (ˇ) and the circumflex (ˆ) to deal with these
problems: the caron changes \emph{Níkẹ̀ẹ́} to ``Níkẹ̌'' while the
circumflex restores the contour to ``Fâjuyì'' and eliminates the
ambiguity.

The notation proposed here has formal precedent in the phonological
literature. Olmsted (1951), in the earliest phonemic analysis of Yorùbá
published in an international journal, describes nine pitch phonemes:
three level and six contour tones, and employs the caron and circumflex
as symbols for low-high rising and high-low falling realizations
respectively. Oyětádé (1988) employs the same symbols in a full
autosegmental analysis of Yorùbá tone. Both works, however, confine the
notation to phonetic and phonological transcription. The present proposal advances the symbols as orthographic conventions, accessible through standard keyboard input tools and integrated into a computational synthesis pipeline, and proposes them as a revision of the standard Yorùbá orthography: an addition to its inventory of tone marks, rather than a restructuring of the system.

\section{2. The Problem: Contour Tones Without Host
Syllables}\label{the-problem-contour-tones-without-host-syllables}

\subsection{2.1 Phonological background}\label{phonological-background}

In standard Yorùbá orthography, a vowel letter carries at most one tone
diacritic. When a contour tone must be represented, the convention is
vowel gemination: the vowel is written twice, each instance bearing one
of the two component tones. Thus the name ``Níkẹ̀ẹ́'', already discussed
above, and others like Ọmọ́tọ́ṣọ̀ọ́, Ajéìígbé, or Oyèélànà.\footnote{In
  practice, these geminated forms are rarely used; writers typically
  revert to the shorter unmarked spelling, with single vowels.} The
representation is phonetically accurate but visually unfamiliar to
speakers accustomed to the shorter forms like Níkẹ, Ọmọ́tọ́ṣọ, Ajéigbé, or
Oyèlànà.\footnote{See also Ọlátúbọ̀sún 2012, pp. 13--15, for experimental
  data on the importance of rising and falling tones in the assessment
  of naturalness in non-native speakers.}

There are two types of contour tone in Yorùbá. The first is a
\emph{phonetic contour}, arising from tonal assimilation across adjacent
syllables: when a high-toned syllable is immediately followed by a
low-toned syllable, the transition between them produces a falling
percept at the boundary, and vice versa. A name like \emph{Táyọ̀}, for
instance, exhibits a falling quality on the final syllable produced by
the juxtaposition of the preceding high tone \emph{á} with the lexically
low \emph{ọ̀}. This contour is already fully and correctly represented by
the existing orthographic system, with the same acute/grave accents
fully representing the High/Low tone manifestations when contour is
absent, and the Rising/Falling tone when it is present.\footnote{This is
  why the convention is not to write the name instead as ``Táyộ''.} This
type of phonetically conditioned F0 movement at tonal boundaries is well
documented in Yorùbá (Connell \& Ladd, 1990).

The second type is a \emph{lexical contour}, arising when two distinct
tonal specifications must be realized within a single syllable. This
occurs most visibly in names and words whose phonological history
involves vowel elision or historical compounding, leaving two tones
compressed onto one vowel. This paper is focused on this second type:
the geminated contour tones on single vowel positions, with
recommendations for the caron and circumflex convention use.

\subsection{2.2 The corpus of affected
names}\label{the-corpus-of-affected-names}

Personal names are especially susceptible to this problem for several
reasons. Names, for a start, often encode compressed semantic content
derived from longer phrases or proverbs, making contour tones
phonologically expected. Secondly, they acquire stable social identities
through repeated use in documents, signage, and official records where
tone marks are commonly omitted altogether. Third, bearers of names
frequently (and often obstinately) resist alterations to established
spellings even when offered phonologically superior alternatives. 

The author first publicly identified this problem in 2019, in a translator’s note to his Yorùbá translation of Chimamanda Ngozi Adichie’s “The Shivering” (Túbọ̀sún, 2019). Using the name Adenikẹ (Table 1, entry 3), the note showed that no existing spelling is both lexically and phonologically faithful, and suggested the caron or circumflex as a solution. It also recorded that the input tool then available could not place a caron on a sub-dotted vowel. The problem was raised again in a presentation at the British Library in 2020 (Túbọ̀sún, 2020), where personal names with unrepresentable contour tones were listed as an unresolved orthographic challenge.

The following \emph{Table 1} illustrates twenty representative names and
words, showing their natural spelling (without tone marks), common
spelling (with improvised tone marks of acute or grave accents), the
fully-expanded tonal form using vowel gemination, and the proposed
contour-marked compact form.

Entries 1--15 represent names where the contour is absent from
conventional spelling, while entries 16--20 represent a slightly
different set of words where the full tonal form exists but is
typographically awkward. Seven of these entries (1–6 and 8) also appear, with their synthesis file keys, in Túbọ̀sún, Olúòkun, et al. (2026, Table 5); the present table extends that sample to the full set.
Words like \emph{Káàárọ̀} ("good morning"),
\emph{Olóòórùn} ("unhygienic person"), \emph{Olóòótọ́} ("a truthful
person"), \emph{Aláàáké} ("an axe wielder"), and \emph{Aláàárẹ̀} ("a sick
person") carry contour tones whose full tonal forms are both
semantically necessary but visually impractical. The three contiguous
vowels make them typographically awkward in most manual and digital
contexts, leading writers to use shortened forms that misrepresent or
obscure the tonal structure. 

The caron resolves both problems simultaneously: \emph{Káǎrọ̀},
\emph{Olóǒrùn}, \emph{Olóǒtọ́}, \emph{Aláǎké}, \emph{Aláǎrẹ̀} preserve the
semantic content of the contour while eliminating the triple-vowel
sequence.

\begin{longtable}{@{}
  >{\raggedright\arraybackslash}p{(\linewidth - 8\tabcolsep) * \real{0.0510}}
  >{\raggedright\arraybackslash}p{(\linewidth - 8\tabcolsep) * \real{0.3192}}
  >{\raggedright\arraybackslash}p{(\linewidth - 8\tabcolsep) * \real{0.1983}}
  >{\raggedright\arraybackslash}p{(\linewidth - 8\tabcolsep) * \real{0.2318}}
  >{\raggedright\arraybackslash}p{(\linewidth - 8\tabcolsep) * \real{0.1997}}@{}}
\caption{Natural form, common (improvised) spellings, full tonal forms, and
proposed contour-marked forms for personal names and lexical items.}\label{tab:contours}\\
\toprule
 & \textbf{Name/Word (Common spelling with no tone marks)}
 & \textbf{Common spelling workaround (omitting the contour)}\footnotemark
 & \textbf{Full phonologically accurate written form}
 & \textbf{Contour-marked form (Proposed)} \\
\midrule
\endfirsthead
\toprule
 & \textbf{Name/Word (Common spelling with no tone marks)}
 & \textbf{Common spelling workaround (omitting the contour)}
 & \textbf{Full phonologically accurate written form}
 & \textbf{Contour-marked form (Proposed)} \\
\midrule
\endhead
\bottomrule
\endlastfoot
1.  & Aromọlaran (personal name)         & \emph{Arómọlárán} & Arómọláràán & Arómọlárǎn \\ \addlinespace
2.  & Ijẹrisi (\emph{testimony})         & \emph{Ìjẹ́risí}    & Ìjẹ́rìísí    & Ìjẹ́rǐsí \\ \addlinespace
3.  & Adenikẹ (personal name)            & \emph{Adéníkẹ́}    & Adéníìkẹ́    & Adénîkẹ́ \\ \addlinespace
4.  & Ajafẹtọ (\emph{an advocate})       & \emph{Ajàfẹ́tọ}    & Ajàfẹ́tọ̀ọ́    & Ajàfẹ́tọ̌ \\ \addlinespace
5.  & Adebiyi (personal name)            & \emph{Adébíyí}    & Adébíyìí    & Adébíyǐ \\ \addlinespace
6.  & Alawiye (\emph{the explainer})     & \emph{Aláwiyé}    & Aláwìíyé    & Aláwǐyé \\ \addlinespace
7.  & Onikan (place name)                & \emph{Oníkán}     & Oníkàán     & Oníkǎn \\ \addlinespace
8.  & Oyedeji (personal name)            & \emph{Oyèdéjì}    & Oyèédèjì    & Oyědèjì \\ \addlinespace
9.  & Muyiwa (personal name)             & \emph{Múyìwá}     & Múyìíwá     & Múyǐwá \\ \addlinespace
10. & Ọmọtọṣọ (personal name)            & \emph{Ọmọ́tọ́ṣọ}    & Ọmọ́tọ́ṣọ̀ọ́    & Ọmọ́tọ́ṣọ̌ \\ \addlinespace
11. & Ayọdeji (personal name)            & \emph{Ayọ̀déjì}    & Ayọ̀ọ́dèjì    & Ayọ̌dèjì \\ \addlinespace
12. & Ikorira (\emph{hatred/spite})      & \emph{Ìkóríra}    & Ìkórìíra    & Ìkórǐra \\ \addlinespace
13. & Akẹkọ (\emph{student})             & \emph{Akẹ́kọ̀}      & Akẹ́kọ̀ọ́      & Akẹ́kọ̌ \\ \addlinespace
14. & Lakọkọ (\emph{firstly})            & \emph{Lákọkọ́}     & Lákọ̀ọ́kọ́     & Lákọ̌kọ́ \\ \addlinespace
15. & Sufe (\emph{whistle})              & \emph{Súfé}       & Súfèé       & Súfě \\ \addlinespace
16. & Kaarọ (\emph{good morning})        & \emph{Káàrọ̀}      & Káàárọ̀      & Káǎrọ̀ \\ \addlinespace
17. & Oloorun (\emph{Unhygienic person}) & \emph{Olóòrùn}    & Olóòórùn    & Olóǒrùn \\ \addlinespace
18. & Olooto (\emph{a truthful person})  & \emph{Olóòtọ́}     & Olóòótọ́     & Olóǒtọ́ \\ \addlinespace
19. & Alaake (\emph{an axe wielder})     & \emph{Aláàké}     & Aláàáké     & Aláǎké \\ \addlinespace
20. & Alaare (\emph{a sick person})      & \emph{Aláàrẹ̀}     & Aláàárẹ̀     & Aláǎrẹ̀ \\
\end{longtable}
\footnotetext{The work-around, which involves using an acute or grave accent
  to mark the contour, or merely leaving it blank, renders the pronunciation
  inaccurate, so none of these entries would generate accurate intelligible
  pronunciations as marked, as can be verified with the synthesizer at
  \href{http://www.ttsyoruba.com}{\ul{www.TTSYoruba.com}}.}

In \emph{Table 2}, we see a related but more acute category of cases,
where the conventional spelling converts the name's meaning entirely,
sometimes in the opposite direction. Several common Yorùbá names that
derive from the formula \emph{Ifá + negative verb}, for instance, fall
into this category, where the falling contour on the initial syllable
carries the negation. Writing with a simple high tone instead, which
current convention and keyboard limitations mandated, obviates the
negative meaning and stipulates the opposite of what was intended.
\emph{Ifâpurọ́}, for example, means ``Ifá did not lie''; written as
\emph{Ifápurọ́}, however, it means ``Ifá lies'', while \emph{Fâjuyìgbé}
means ``Ifá did not let honour go to waste'' while \emph{Fájuyìgbé} (or
\emph{Fájuyì}, as it is commonly written), means the
opposite.\footnote{The circumflex is offered here on this table as an
  option for writers and publishers who wish to restore semantic
  accuracy, not as a prescription that all bearers of these names must
  alter their established spellings. A bearer of the name \emph{Ifápurọ́}
  who has used that form throughout their life remains free to continue
  doing so. See Section~5 for the recommended use.}

{\small
\begin{longtable}{@{}
  >{\raggedright\arraybackslash}p{(\linewidth - 8\tabcolsep) * \real{0.0500}}
  >{\raggedright\arraybackslash}p{(\linewidth - 8\tabcolsep) * \real{0.3300}}
  >{\raggedright\arraybackslash}p{(\linewidth - 8\tabcolsep) * \real{0.1900}}
  >{\raggedright\arraybackslash}p{(\linewidth - 8\tabcolsep) * \real{0.1900}}
  >{\raggedright\arraybackslash}p{(\linewidth - 8\tabcolsep) * \real{0.2400}}@{}}
\caption{Names where conventional tone marking produces an opposite or
unintended meaning. The common spelling column shows the name as currently
written and the meaning that results. The proposed contour form uses the
circumflex (â) to mark the falling H+L contour on the syllable bearing the
negative contour, restoring the negative meaning encoded in the name's
etymology.}\label{tab:inverted}\\
\toprule
 & \textbf{Common spelling (unintended meaning)}
 & \textbf{Full geminated form (rarely to never used)}
 & \textbf{Proposed contour form (with circumflex)}
 & \textbf{Correct meaning} \\
\midrule
\endfirsthead
\toprule
 & \textbf{Common spelling (unintended meaning)}
 & \textbf{Full geminated form (rarely to never used)}
 & \textbf{Proposed contour form (with circumflex)}
 & \textbf{Correct meaning} \\
\midrule
\endhead
\bottomrule
\endlastfoot
1.  & Ifápurọ́ (``Ifá lies'') & Ifáàpurọ́ & Ifâpurọ́ & Ifá did not lie. \\ \addlinespace
2.  & Fájuyìgbé / Fájuyì (``Ifá lets honour go to waste'') & Fáàjuyìgbé / Fáàjuyì & Fâjuyìgbé / Fâjuyì & Ifá did not let honour go to waste. \\ \addlinespace
3.  & Adéjímitẹ́ (``Royalty let me be shamed'') & Adéèjímitẹ́ & Adêjímitẹ́ & Royalty prevented my shame. \\ \addlinespace
4.  & Ifátànmí (``Ifá deceived me'') & Ifáàtànmí & Ifâtànmí & Ifá did not deceive me. \\ \addlinespace
5.  & Fágbàmígbé (``Ifá forgot me'') & Fáàgbàmígbé & Fâgbàmígbé & Ifá did not forget me. \\ \addlinespace
6.  & Fásọ̀ràntì (``Ifá failed to settle the rift'') & Fáàsọ̀ràntì & Fâsọ̀ràntì & Ifá did not fail to settle the rift. \\ \addlinespace
7.  & Kúpolúyì (``Death took the honourable one'') & Kúùpolúyì & Kûpolúyì & Death did not take the honourable one. \\ \addlinespace
8.  & Fádojú(tìmí) (``Ifá disappointed me'') & Fáàdójútìmí & Fâdójútìmí & Ifá did not disappoint me. \\ \addlinespace
9.  & Akínṣọ̀tẹ̀ (``The valiant one rebelled'') & Akíìnṣọ̀tẹ̀ & Akînṣọ̀tẹ̀ & The valiant one did not rebel. \\ \addlinespace
10. & Ọláyìnmínù (``Honour ignored me'') & Ọláàyìnmínù & Ọlâyìnmínù & Honour did not ignore me. \\
\end{longtable}
}

\subsection{2.3 Prior practice and the handwriting
record}\label{prior-practice-and-the-handwriting-record}

The problem described here was not invisible to earlier Yorùbá writers.
Earlier writing forms had deployed the tilde (\textasciitilde) to mark
contour. But this was recommended for discontinuation by the 1974 Yorùbá
Orthography Committee, whose recommendations drew on the grammatical and
orthographic tradition established by Bámgbóṣé (1966) and remain the
authoritative standard to date (Joint Consultative Committee on
Education, 1974; Awóbùlúyì, 1994). Since then, informal handwriting
traditions in southwestern Nigeria have employed improvised wedge-shaped
marks combining the visual forms of the acute and grave diacritics to
signal contour tones. These marks, resembling the characters {[}v{]}
(for rising) and {[}\raisebox{-0.1em}{\includegraphics[height=0.8em]{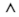}}{]} (for falling), were widespread enough to be
intelligible among literate Yorùbá speakers, but were never officially
codified in subsequent orthographic recommendation that the author has
been able to locate.

Ògúnbíyǐ (2003), in his study of Yoruba \emph{Ajami} writing, which
predates the Roman orthographic tradition, documents that tones were ``by far the greatest problem to which the new standardized anjemi has no solutions yet''; a failure that persisted across two centuries of
attempts in both the Arabic-script and Roman-script traditions (p. 97).
The contour tone problem is therefore not a recent computational
inconvenience but a structural gap in every writing system that has been
applied to Yorùbá.

\section{3. Proposed Solution: Caron and Circumflex as Contour
Markers}\label{proposed-solution-caron-and-circumflex-as-contour-markers}

\subsection{3.1 Character selection}\label{character-selection}

We propose the following convention:

\begin{itemize}
\item
  Caron (ˇ), yielding e.g. ǎ, ě, ẹ̌, ǐ, ǒ, ọ̌, ǔ: to encode a rising (L+H)
  contour on a single vowel.
\item
  Circumflex (ˆ), yielding e.g. â, ê, ệ, î, ô, ộ, û: to encode a falling
  (H+L) contour on a single vowel.
\end{itemize}

These characters are already defined in Unicode and supported in most
modern fonts, including those commonly used in Yorùbá digital
publishing. They are visually distinguishable from one another and from
the three primary Yorùbá tone diacritics (acute, grave,
macron/unmarked). Crucially, they can also be combined with the
under-dot subscript used for the open vowels \emph{ẹ} and \emph{ọ} which
are among the most common sites of contour tone in Yorùbá.

This means the full proposed diacritic inventory for contour-bearing
open vowels is: \emph{ệ} (falling on open-e), \emph{ẹ̌} (rising on
open-e), \emph{ộ} (falling on open-o), \emph{ọ̌} (rising on open-o), in
addition to the standard vowels with caron or circumflex.\footnote{While
  placing a tone diacritic above an under-dot vowel can occasionally
  cause vertical clipping in some legacy fonts or restricted mobile
  environments, modern OpenType fonts and Unicode Normalization Form C
  (NFC) rendering ensure stable and legible display in most current
  digital environments, though legacy systems and restricted mobile
  rendering contexts may still exhibit clipping.}

\subsection{3.2 Equivalence with geminated
forms}\label{equivalence-with-geminated-forms}

The proposed marks are not intended to replace the geminated spelling
where it is already established and accepted. Rather, they are offered
as an orthographic alternative for contexts where the conventional
doubled-vowel form is unavailable or undesired.\footnote{Because Olmsted
  (1951) already recommended the caron for rising contour and circumflex
  for falling contour for use in phonetic notations, the substantive
  innovation here is their recommended orthographic use. } A key design
principle is therefore equivalence: the contour-marked form and the
geminated form should be treated as phonologically identical by any
computational system processing Yorùbá text.

Formally, the following equivalences hold:

\begin{quote}
ǎ $\equiv$ àá (rising contour on /a/)

â $\equiv$ áà (falling contour on /a/)

ě $\equiv$ èé (rising contour on /e/)

ê $\equiv$ éè (falling contour on /e/)

ẹ̌ $\equiv$ ẹ̀ẹ́ (rising contour on /ẹ/)

ệ $\equiv$ ẹ́ẹ̀ (falling contour on /ẹ/)

ǐ $\equiv$ ìí (rising contour on /i/)

î $\equiv$ íì (falling contour on /i/)

ǒ $\equiv$ òó (rising contour on /o/)

ô $\equiv$ óò (falling contour on /o/)

ọ̌ $\equiv$ ọ̀ọ́ (rising contour on /ọ/)

ộ $\equiv$ ọ́ọ̀ (falling contour on /ọ/)

ǔ $\equiv$ ùú (rising contour on /u/)

û $\equiv$ úù (falling contour on /u/)
\end{quote}

These equivalences must be explicitly encoded in any lexicon, text
normalizer, or speech synthesis pipeline that processes Yorùbá
text.\footnote{Furthermore, to prevent database fragmentation and ensure
  discoverability, information retrieval systems and search algorithms
  should be programmed to treat these contour-marked forms as
  semantically and structurally identical to their geminated
  counterparts during text indexing and querying.}

\subsection{3.3 Beyond personal names: contours in English loanwords}\label{3.3 Beyond personal names: contours in English loanwords}

The falling contour also arises productively in the adaptation of loanwords. When the stressed syllable of an English word is heavy, whether it contains a long vowel or is closed by a consonant, its vowel typically surfaces in Yorùbá with a falling (H+L) contour before an epenthetic final vowel: \textit{code} → kôdù, \textit{phone} → fônù, \textit{shop} → ṣộbù, \textit{film} → fîmù, \textit{tablet} → tábúlệtì. On the mapping of English stress to Yorùbá tone more generally, see Kenstowicz (2006). Current practice writes these with a doubled vowel (kóòdù, fóònù, ṣọ́ọ̀bù), and the hand-verified ÌròyìnSpeech text corpus contains many such spellings (e.g. bọ́ọ̀lù ‘ball’, fóònù ‘phone’, káàdì ‘card’, nọ́ọ̀sì ‘nurse’; Túbọ̀sún, Àrẹ̀mú, et al., 2026). In that corpus, 65 English-derived words with a single-quality falling contour occur 840 times, every one written with a doubled vowel (Table 3). English diphthongs, by contrast, are adapted as sequences of two distinct vowels (\textit{file} → fáìlì, \textit{kite} → káìtì) and are correctly written with two letters. The circumflex is therefore proposed only where a single vowel quality carries the contour.

\begin{table}[h]
\centering
\small
\begin{tabular}{lllr}
\hline
English & Current spelling & Proposed form & Tokens \\
\hline
ball     & bọ́ọ̀lù\textsuperscript{a} & bộlù    & 476 \\
phone    & fóònù      & fônù     & 34 \\
goal     & góòlù      & gôlù     & 28 \\
card     & káàdì      & kâdì     & 23 \\
church   & ṣọ́ọ̀ṣì     & ṣộṣì     & 22 \\
nurse    & nọ́ọ̀sì     & nộsì     & 21 \\
gas      & gáàsì      & gâsì     & 18 \\
shop     & ṣọ́ọ̀bù     & ṣộbù     & 18 \\
film     & fíìmù      & fîmù     & 15 \\
court    & kọ́ọ̀tù     & kộtù     & 13 \\
class    & kíláàsì    & kílâsì   & 13 \\
office   & ọ́fíìsì    & ọ́fîsì   & 11 \\
\hline
\multicolumn{3}{l}{All 65 English loanwords of this type} & 840 \\
\hline
\end{tabular}
\caption{English loanwords with a falling contour on a single vowel quality:
current geminated spellings and their frequency in the ÌròyìnSpeech text
corpus (Túbọ̀sún, Aremu, et al., 2026), with proposed circumflex forms.
\textsuperscript{a}Includes compounds such as \emph{agbábọ́ọ̀lù}
`footballer'.}
\label{tab:loanwords}
\end{table}

\section{4. Computational
Implementation}\label{computational-implementation}

\subsection{4.1 Keyboard input}\label{keyboard-input}

The WriteYoruba keyboard software (available for macOS and Windows at
WriteYoruba.com, first released 2016) was extended to include caron and
circumflex key combinations covering all Yorùbá vowels, including the
open variants ẹ and ọ. On the WriteYoruba keyboard, the caron is typed with Option + comma (AltGr + comma on Windows) and the circumflex with Option + period (AltGr + period). A platform-neutral reference mapping, with the full character table and normalization requirements, is available at Túbọ̀sún, Le Guen, et al. (2026).\footnote{Openly available on
Zenodo under CC BY 4.0: \url{https://doi.org/10.5281/zenodo.23305200}.}

This extension addresses a gap left by existing major-platform keyboard
tools. The Google Gboard keyboard, which supports Yorùbá and was
designed with input from a team that included the author, provides standard level-tone diacritics across all seven Yorùbá vowels, including the open vowels ẹ and ọ, and some caron and circumflex forms on the plain vowels, but cannot combine these marks with the subdot, so contour-marked open vowels such as ẹ̌, ọ̌, ệ and ộ are unavailable (van Esch et al. 2019). The WriteYoruba extension described here is, to our knowledge, the first Yorùbá keyboard to make the full set of contour-marked vowels, including ẹ̌ and ọ̌, accessible, precisely because the orthographic proposal in Section~3 now provides the standard
they require.

We route these contour mark diacritics through the existing WriteYoruba
infrastructure rather than proposing standalone Unicode input methods
because Yorùbá writers are most likely to adopt new orthographic
conventions if they can be accessed through tools already integrated
into their workflow, as this already is. The keyboard has been used by
students, teachers, and publishers alike since its release in 2016.
There is a broader culture of tone-mark omission that pervades
contemporary Yorùbá writing across domains from journalism to religious
publishing to digital signage, where even level-tone diacritics are
routinely treated as optional (Olúmúyǐwá 2013). This combined with
challenges with Unicode creation of pre-composed characters for African
language letters (see Túbọ̀sún, \emph{2026b}) has made the
challenge more acute and a solution more expedient.

\subsection{4.2 Speech synthesis}\label{speech-synthesis}

The equivalences in Section~3.2 have been implemented in TTSYoruba.com, the browser-based Yorùbá text-to-speech engine developed at YorubaName.com. A normalization step expands caron- and circumflex-marked vowels into their geminated equivalents before grapheme-to-phoneme conversion, so that a contour-marked form such as Arómọlárǎn is synthesized identically to its full tonal form Arómọláràán. The system architecture, the normalization pipeline, and the current state of caron and circumflex support are documented in Túbọ̀sún, Olúòkun, et al. (2026); the complete grapheme-to-audio mapping specification is deposited in Túbọ̀sún (2026a).

That work also reports a listener evaluation (N = 50) comparing synthesized output for matched pairs of geminated and contour-marked names. No significant difference was found between the two notations in either naturalness or intelligibility, providing perceptual support for the equivalence principle set out in Section~3.2. The present paper does not repeat the implementation or evaluation, and is concerned instead with the orthographic case for the notation.

\section{5. Discussion}\label{discussion}

The proposal described here is modest in scope but has implications for
several areas of Yorùbá language technology and orthographic policy.

For computational linguistics, the equivalence mappings in Section~3.2
suggest a straightforward normalization protocol that could be adopted
by any NLP pipeline working with Yorùbá text. Named-entity recognition
systems, text-to-speech engines, and machine translation models that
process Yorùbá names would benefit from consistent handling of contour
tone representations.

For orthographic standardization, the caron/circumflex convention is proposed as a revision of the Yorùbá orthography established by the 1974 Orthography Committee. The revision is additive: it introduces two marks for a category the standard does not currently represent, and leaves the existing tone marks and geminated spellings intact. In this it reverses, with Unicode-compatible characters, the Committee’s removal of the tilde, reinstating single-symbol contour notation that is unambiguous, typeable, and combinable with the existing diacritic system.

For speakers and name-bearers, the proposal offers a resolution to what
has been described anecdotally as a longstanding frustration: the
inability to mark one\textquotesingle s own name correctly on a computer
without either distorting its appearance or leaving the tone
representation incomplete. The contour marks allow full tonal
specification within the familiar one-vowel-per-syllable visual frame.

The wedge-shaped handwriting marks described by some Yorùbá writers are
best understood as informal adaptations of a notation that has possessed
formal phonological standing since Olmsted (1951) and Oyětádé (1988).
What was historically missing was a viable path to bring this
transcription standard into everyday orthography and computer input.
This is a gap that this paper\textquotesingle s computational
implementation now addresses.

\textbf{Recommendation.} The standard Yorùbá orthography, as set out by the 1974 Orthography Committee, provides no means of representing a lexical contour tone on a single vowel; its only resource, vowel gemination, is unavailable wherever spellings are conventionally fixed, as they are for most personal names. We therefore propose that the standard orthography be revised to include the caron and circumflex as the means of marking rising and falling lexical contours on a single vowel. Geminated spellings remain valid, and the two forms are equivalent (Section~3.2). We recommend their use in lexicography, where dictionary headwords and pronunciation fields require full tonal specification; in linguistic, pedagogical and reference writing; and in computational resources, where the equivalence mappings in Section~3.2 allow contour-marked and geminated forms to be treated as identical. Individual name-bearers remain free to keep their established spellings (see note 7). We invite Yorùbá orthographic and language bodies to consider this revision for formal adoption.

The adoption of these recommendations depends on accessibility, visibility, and institutional endorsement. To measure this uptake empirically, future development will track the organic submission of caron and circumflex characters in server-side logs on TTSYoruba.com, alongside opt-in, anonymized telemetry for contour key usage within WriteYoruba. The present implementation, embedded in a keyboard input
tool and tested in a public-facing TTS engine (Túbọ̀sún, Olúòkun, et al., 2026), is intended as a proof of concept that can support broader advocacy for formal inclusion of
contour diacritics in Yorùbá digital typography standards. The decision
to implement the convention through keyboard input and synthesis tools
rather than seeking new Unicode encodings reflects both the practical
timeline (Unicode encoding proposals can take years to materialize), and
the structural barriers documented in Túbọ̀sún (2026b). 

\section{6. Conclusion}\label{conclusion}

Contour tones in Yorùbá personal names represent a recognized but
underaddressed problem in both orthographic theory and computational
practice. While the existing solution of vowel gemination is
phonologically correct, it remains socially unavailable for a
significant corpus of names whose spellings are conventionally fixed. To
resolve this, this paper has proposed a diacritical convention using the
Unicode caron and circumflex on single vowels to mark rising and falling
contour tones respectively. By formalizing the equivalence between
contour marks and their geminated forms, and demonstrating an initial
implementation in keyboard input and speech synthesis software, this
work bridges the historical gap between linguistic transcription and
modern digital typography. The barriers to encoding these characters in
Unicode are documented in Túbọ̀sún (2026b). While institutional
orthographic reform has better reach, this technological implementation
provides an immediate tool for writers and developers. The proposal
invites active engagement from Yorùbá linguists, orthographic
committees, and language technology developers. Future work will build
on this foundation through a systematic corpus analysis of contour-tone
names in the YorubaName.com database, user testing of the keyboard
extension, and further evaluation of TTS output for falling-contour (circumflex) forms on open vowels. 

Further gaps remain in the orthography. Sequences involving the mid tone, as in names conventionally written \textit{Tinubu}, \textit{Oninure}, and \textit{Oladipo} (fully \textit{Tinúubú}, \textit{Onínúure}, and \textit{Ọládiípọ̀)}, likewise have no compact written form and are rarely spelled out in full. Because the mid tone does not form lexical contours (Section~1), these are better analysed as vowel sequences than as contours, and they call for a separate convention. The tilde, removed by the 1974 Orthography Committee, is one candidate, though its established use for nasalization in other orthographies would need to be addressed. A future paper will take up this class of names.

\section{Acknowledgments}\label{acknowledgments}

The author thanks the users of YorubaName.com and WriteYoruba whose
questions and frustrations over many years clarified the dimensions of
the problem described here; and Professor Ronald P. Schaefer, Professor
Karin Barber, and Dr. Túndé Adégbọlá for helpful input.

\section*{Competing interests}
The author founded YorubaName.com, which develops and distributes the \mbox{WriteYoruba} keyboard.

\section{References}\label{references}

Akinlabí, A. (1985). Tonal underspecification and Yorùbá tone. Ph.D.
dissertation, University of Ibadan.

Awóbùlúyì, Q. (1994). The development of Standard Yoruba. In I. Fodor \&
C. Hagège (Eds.), \emph{Language reform: History and future} (Vol. VI,
pp. 25--42). Helmut Buske Verlag.

Bámgbóṣé, A. (1966). A grammar of Yoruba. Cambridge University Press.

Capo, H. B. C. (1989). Defoid. In J. Bendor-Samuel (Ed.), The
Niger-Congo languages (pp. 275--290). University Press of America.

Connell, B., \& Ladd, D. R. (1990). Aspects of pitch realisation in
Yoruba. \emph{Phonology}, \emph{7}(1), 1--29.
\href{https://doi.org/10.1017/S095267570000110X}{\ul{https://doi.org/10.1017/S095267570000110X}}

van Esch, D., Sarbar, E., Lucassen, T., O\textquotesingle Brien, J.,
Breiner, T., Prasad, M., Crew, E., Nguyen, C., \& Beaufays, F. (2019).
\emph{Writing across the world\textquotesingle s languages: Deep
internationalization for Gboard, the Google keyboard.} arXiv preprint
arXiv:1912.01218.
\href{https://arxiv.org/abs/1912.01218}{\ul{https://arxiv.org/abs/1912.01218}}

Joint Consultative Committee on Education. (1974). \emph{Yorùbá
orthography (Àkọtọ́ èdè Yorùbá)}. Federal Ministry of Education, Lagos.

Kenstowicz, M. (2006). Tone loans: The adaptation of English loanwords into Yoruba. In J. Mugane, J. P. Hutchison, \& D. A. Worman (Eds.), \textit{Selected proceedings of the 35th Annual Conference on African Linguistics} (pp. 136–146). Cascadilla Proceedings Project.

Ògúnbíyǐ, I.A. (2003). The search for a Yoruba orthography
since the 1840s: Obstacles to the choice of the Arabic script.
\emph{Sudanic Africa}, \emph{14}, 77--102.

Ọlátúbọ̀sún, K. (2012). \emph{Studies of initial tonal acquisition by
American English speakers learning Yoruba} {[}Master\textquotesingle s
thesis, Southern Illinois University Edwardsville{]}. Zenodo.
\href{https://doi.org/10.5281/zenodo.20833023}{\ul{https://doi.org/10.5281/zenodo.20833023}}

Olmsted, D. L. (1951). The phonemes of Yoruba. \emph{WORD}, \emph{7}(3),
245--249.
\href{https://doi.org/10.1080/00437956.1951.11659409}{\ul{https://doi.org/10.1080/00437956.1951.11659409}}

Olúmúyǐwá, T. (2013). Yoruba writing: Standards and trends.
\emph{Journal of Arts and Humanities}, \emph{2}(1), 40--51.
\href{https://doi.org/10.18533/journal.v2i1.50}{\ul{https://doi.org/10.18533/journal.v2i1.50}}

Oyěláràn, O. (1973). Yorùbá phonology. Ph.D. dissertation, Stanford
University.

Oyětádé, B. A. (1988). \emph{Issues in the analysis of Yorùbá tone}
{[}Doctoral dissertation, University of London{]}. ProQuest
Dissertations \& Theses. (ProQuest No. 10672612).

Pulleyblank, D. (2004). A note on tonal markedness in Yoruba.
\emph{Phonology}, \emph{21}(3), 409--425.
\href{https://doi.org/10.1017/S0952675704000326}{\ul{https://doi.org/10.1017/S0952675704000326}}

Túbọ̀sún, K. (2016). WriteYoruba: A keyboard input system for Yorùbá and
Igbo. YorubaName.com.
\href{https://writeyoruba.com}{\ul{https://writeyoruba.com}}

Túbọ̀sún, K. (2019). The shivering, by Chimamanda Ngozi Adichie [Translation, with translator’s note]. \textit{Absinthe: World Literature in Translation}, \textit{26}. \href{https://doi.org/10.3998/absinthe.9518}{https://doi.org/10.3998/absinthe.9518}

Túbọ̀sún, K. (2020, September 2). \emph{Yorùbá orthography from Àjàyí
Crowther to date: Through the collection items at the British Library:
Progress and problems} {[}Conference presentation{]}.
\textquotesingle How Should We Write Yorùbá?\textquotesingle{} British
Library Webinar, London.
\href{https://www.youtube.com/watch?v=HDn2ou7umxs}{\ul{https://www.youtube.com/watch?v=HDn2ou7umxs}}

Túbọ̀sún, K. (2026a). \emph{A grapheme-to-audio mapping and phonological
rule system for Yorùbá concatenative speech synthesis}. Zenodo.
\href{https://doi.org/10.5281/zenodo.21500364}{\ul{https://doi.org/10.5281/zenodo.21500364}}

Túbọ̀sún, K. (2026b). Yorùbá in Unicode: An overview of a
problem. \href{arXiv:2609.33734).} arXiv. {\ul{https://doi.org/10.48550/arXiv.2609.33734}} (To appear in \textit{Yorùbá print culture: A handbook}, Routledge)

Túbọ̀sún, K., Aremu, A., Ògúnrẹ̀mí, T., Orife, I., \& Adelani, D. I. (2026). \textit{The ÌròyìnSpeech text corpus: 24,905 curated Yorùbá sentences for speech and language technology}. arXiv preprint https://arxiv.org/abs/2610.05366v4 

Túbọ̀sún, K., Le Guen, L., Morais, L., \& Williamson, V. (2026). \textit{Yorùbá contour-tone vowels: A reference keyboard mapping} (Version 1.0) [Data set]. Zenodo. \href{https://doi.org/10.5281/zenodo.23305200}{\ul{https://doi.org/10.5281/zenodo.23305200}}

Túbọ̀sún, K., Olúòkun, A., Adéwuyì, H., \& Adérẹ̀mí, D. (2026). A
situational speech synthesizer for Yorùbá: System design, phonological
rule architecture, and orthographic extensions for contour tones. arXiv
preprint arXiv:2607.18317.
\href{https://arxiv.org/abs/2607.18317}{\ul{https://arxiv.org/abs/2607.18317}}

\end{document}